\documentclass[lettersize,journal]{IEEEtran}
\usepackage{amsmath,amsfonts}
\usepackage{dsfont}
\usepackage{algorithmic}
\usepackage{algorithm}
\usepackage{array}
\usepackage[caption=false,font=normalsize,labelfont=sf,textfont=sf]{subfig}
\usepackage{textcomp}
\usepackage{stfloats}
\usepackage{url}
\usepackage{verbatim}
\usepackage{graphicx}
\usepackage{cite}
\usepackage{booktabs}
\usepackage{multirow}
\usepackage{arydshln}

\makeatletter
\let\NAT@parse\undefined
\makeatother
\usepackage{hyperref}  
\begin{document}

\title{DPDETR: Decoupled Position Detection Transformer for Infrared-Visible Object Detection
}

\author{Junjie Guo, Chenqiang Gao*, Fangcen Liu and Deyu Meng, \textit{Member, IEEE}
\thanks{$^\ast$ Corresponding author.}
\thanks{Junjie Guo and Fangcen Liu are with the School of Communications and Information Engineering, Chongqing University of Posts and Telecommunications, Chongqing 400065, China.
}
\thanks{Chenqiang Gao is with the School of Intelligent Systems Engineering, Shenzhen Campus of Sun Yat-sen University, Shenzhen, Guangdong 518107, P.R. China. E-mail: gaochq6@mail.sysu.edu.cn.
}
\thanks{Deyu Meng is with the School of Mathematics and Statistics, Xi’an Jiaotong University, Xi’an, Shanxi, 710049, China. E-mail: dymeng@mail.xjtu.edu.cn
}
}

\markboth{June~2025}%
{Shell \MakeLowercase{\textit{et al.}}: A Sample Article Using IEEEtran.cls for IEEE Journals}


\maketitle
\begin{abstract}
Infrared-visible object detection aims to achieve robust object detection by leveraging the complementary information of infrared and visible image pairs. 
However, the commonly existing modality misalignment problem presents two challenges: fusing misalignment complementary features is difficult, and current methods cannot reliably locate objects in both modalities under misalignment conditions.
In this paper, we propose a Decoupled Position Detection Transformer (DPDETR) to address these issues. Specifically, we explicitly define the object category, visible modality position, and infrared modality position to enable the network to learn the intrinsic relationships and output reliably positions of objects in both modalities. To fuse misaligned object features reliably, we propose a Decoupled Position Multispectral Cross-attention module that adaptively samples and aggregates multispectral complementary features with the constraint of infrared and visible reference positions.
Additionally, we design a query-decoupled Multispectral Decoder structure to address the the conflict in feature focus among the three kinds of object information in our task and propose a Decoupled Position Contrastive DeNoising Training strategy to enhance the DPDETR's ability to learn decoupled positions. 
Experiments on DroneVehicle and KAIST datasets demonstrate significant improvements compared to other state-of-the-art methods. The code will be released at \url{https://github.com/gjj45/DPDETR}.
\end{abstract}


\section{Introduction}
\IEEEPARstart{O}{bject} detection is a fundamental task in computer vision, which has been used in various practical applications, e.g., video surveillance, autonomous driving, aerial object detection, etc. With the development of deep learning, the object detection technique has made great progress \cite{ren2016faster,liu2016ssd,redmon2018yolov3}. Since existing methods are mainly designed for visible images, they are still challenged by poor imaging conditions, such as low illumination, smoke, fog, and so on. Thus, infrared images are introduced into the object detection task \cite{herrmann2018cnn,kieu2020task,liu2024infmae}. Unlike visible imaging, infrared imaging cannot capture detailed information but is unaffected by illumination, smoke, and fog occlusion conditions. To achieve robust full-time object detection, infrared-visible object detection by integrating infrared and visible image complementary information has attracted extensive attention in recent years \cite{zhou2023position,zhang2021weakly,tu2022weakly,fu2023lraf,li2023stabilizing,shen2024icafusion,yuan2024improving,yuan2024c}.

\begin{figure}
    \centering
    \includegraphics[trim=60mm 67mm 63mm 72mm,clip,width=0.43\textwidth]{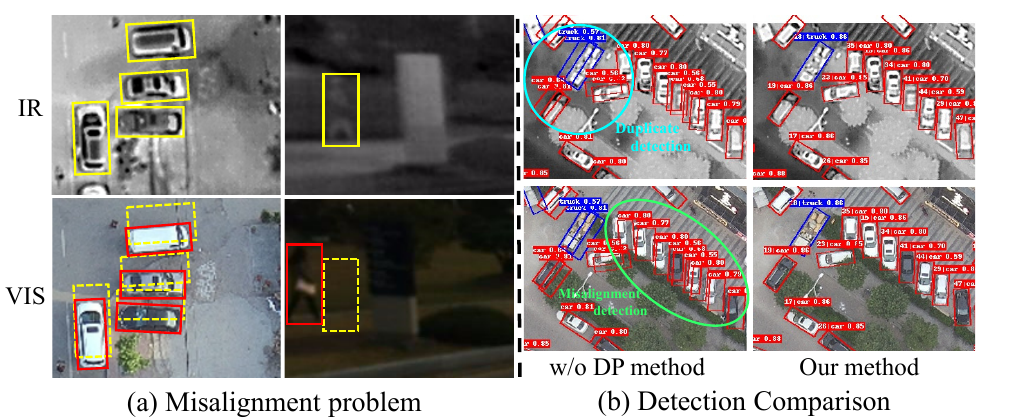}
    \caption{Misalignment problem in infrared-visible object detection and the detection comparison between the common method and our method. (a) Examples of modality misalignment in infrared and visible image pairs. The yellow and red boxes represent the position of the same object in the infrared and visible modalities, respectively. The yellow dashed line represents the object's position in the infrared modality. (b) An example of w/o DP (without Decoupled Position) method where the detection results make it difficult to distinguish the same object in two modalities due to misalignment. Our method can reliably locate objects in both modalities and identify the same object with the same ID. }
    \label{fig:fig1_motivation}
\end{figure}

However, 
most of infrared-visible object detection
methods usually assume that infrared-visible image pairs are well-aligned
, which usually does not hold on in practice. As 
shown in Fig. \ref{fig:fig1_motivation} (a), even through manual registration, precise registration is difficult because infrared-visible images often exhibit significant visual differences and are not always captured at the same timestamp \cite{zhu2023multi}. As a result, the imaging objects in two modalities for the same one are usually misaligned. This could disrupt the consistency of fused feature representation of current methods, 
futher decreasing the detection performance. 

When infrared and visible image pairs are misaligned, the most methods usually could not to obtain reliable object detection results in both modalities.
Even though objects can be correctly located in one modality (reference modality), this still leads to obvious misalignment detection errors in the other modality and even shows confusing duplicate detection results.
As shown in Fig. \ref{fig:fig1_motivation} (b), misalignment detection errors and confusing duplicate detection seriously hinder the recognition of the same object in two modalities, especially when objects are dense and the bounding boxes are tightly constrained.
This phenomenon also indicates that these methods potentially select the modality that best matches the annotated bounding box as the main reference modality, which could lead to unfair treatment of both modalities during feature fusion, resulting in a bias towards the reference modality.
Therefore, it is critical to accurately locate the position of the same object in both infrared and visible modalities.

Recently, a few approaches have tended to address the misalignment problem in infrared-visible object detection. Zhang et al. \cite{zhang2019weakly}, Zhou et al. \cite{zhou2020improving} and Yuan et al. \cite{yuan2022translation,yuan2024improving,yuan2024c} solved the modality misalignment problem by predicting the proposals' offset between two modalities in the RoI head or by predicting the feature points' offset before feature fusion. Although these methods address the alignment of object features to some extent, they 
are unable to align misaligned features in complex scenarios since these methods use a main reference modality proposal to predict the object proposal offset in the other modality directly, without fully utilizing both infrared and visible image features. 
Furthermore, these methods usually output the positions of objects in the reference modality but cannot simultaneously output the reliably positions and one-to-one correspondences of the objects in both modalities. 

In this paper, we propose a novel Decoupled Position Detection Transformer method called DPDETR to achieve instance-level feature alignment and reliably position output of the same object in both modalities. 
Specifically, we decouple the object position in object detection into infrared and visible positions and represent an object with category, infrared position, and visible position information. We explicitly optimize these three aspects for each object by making full use of complementary features. 
To achieve aligned complementary feature fusion, we propose Decoupled Position Multispectral Deformable cross-attention, which performs adaptive sampling and aggregation at the decoupled reference positions of the object in infrared and visible features. Since optimizing these three types of information simultaneously is more complicated and they have conflicting optimization requirements,
we design a query decoupled structure to achieve decoupled cross-attention for fine-optimizing each type of information. Furthermore, we design Decoupled Position Contrastive DeNoising Training to increase the diversity of misalignment situations in the two modalities, helping the network learn and accelerate this decoupled position optimization paradigm.

In summary, our contributions are listed as follows:

\begin{itemize}
\item We propose a novel method called DPDETR with Decoupled Position Multispectral Deformable cross-attention to solve modality misalignment problems and output reliable positions of each object in both modalities. To the best of our knowledge, this is the first work to decouple object positions in infrared and visible modality and simultaneously optimize them using multimodal features.  
\item We design a query decoupled structure to achieve decoupled cross-attention for fine-optimizing each type of information and propose Decoupled Position Contrastive DeNoising Training to assist network training. These two approaches further enhance DPDETR's ability to decouple learning.
\item 
To evaluate the validity of our methods, we tested our DPDETR on both oriented and horizontal infrared-visible object detection tasks.
Extensive experiments on the DroneVehicle dataset and the KAIST dataset show that the proposed method achieves state-of-the-art performance. 
\end{itemize}

\section{Related work}
\subsection{Infrared-visible Object Detection}
Previous research in infrared-visible object detection has mainly relied on one-stage detectors, such as YOLO \cite{redmon2016you,redmon2017yolo9000,redmon2018yolov3}, and two-stage detectors, such as Faster RCNN \cite{ren2016faster}. To utilize the complementary information of infrared and visible images, Wagner et al. \cite{wagner2016multispectral} first constructed the early and late CNN fusion architecture to improve the reliability of object detection. Konig et al. \cite{konig2017fully} introduced a fully convolutional fusion RPN network, which fused features by concatenation, and concluded that halfway fusion can obtain better results \cite{liu2016multispectral}. On this foundation, \cite{cao2023multimodal,qingyun2022cross,roszyk2022adopting} designed CNN-based attention modules to better fuse infrared and visible features. \cite{qingyun2021cross,fu2023lraf,zhu2023multi,shen2024icafusion} introduced transformer-based fusion modules to fuse more global complementary information between infrared and visible images. In addition to directly fusing image features, \cite{li2019illumination,zhou2020improving,yang2022baanet} adopted the illumination-aware fusion method to fuse infrared and visible image features or post-fuse the multibrance detection results. To achieve differential fusion of different regions, \cite{li2018multispectral,cao2019box,li2019illumination,zhang2020multispectral,zhang2021guided} introduced bounding box-level semantic segmentation to guide the fusion of segmented regions, and \cite{kim2021uncertainty} achieved regional-level feature fusion through regions of interest (ROI) prediction. \cite{li2022confidence,li2023stabilizing} further utilized the confidence or uncertainty scores of regions to post-fuse the predictions of multibranchs. However, these methods overlook the modality misalignment problem, resulting in their inability to utilize the misaligned object features. Thus, we propose the novel DPDETR to solve the misalignment problem in infrared-visible object detection.

\subsection{Alignment Learning in Infrared-visible Object Detection}
Modality misalignment is a key problem in infrared-visible object detection. Recently, some work has been dedicated to solving this problem. Zhang et al. \cite{zhang2019weakly,zhang2021weakly} first addressed the alignment problem by predicting the shift offset of the reference proposal in another modality and fusing aligned proposal features. \cite{yuan2022translation,yuan2024improving} further considered the scale and angle offsets of the reference proposal to realize more accurate alignment feature fusion in aerial object detection. \cite{yuan2024c} calculates the attention value between feature points in the reference modality and another modality to achieve the fusion of misaligned object features. However, these methods simply predict the offset of another modality based on the reference modality, without fully utilizing infrared and visible features to learn the intrinsic relationship between the same object in two modalities. In contrast, our method explicitly defines the position of each object in both infrared and visible modalities, fully utilizes the multispectral features to learn this intrinsic relationship, and outputs the reliable position of the object in both modalities.  

\begin{figure*}
    \centering
    \includegraphics[trim=45mm 60mm 45mm 60mm,clip,width=0.95\textwidth]{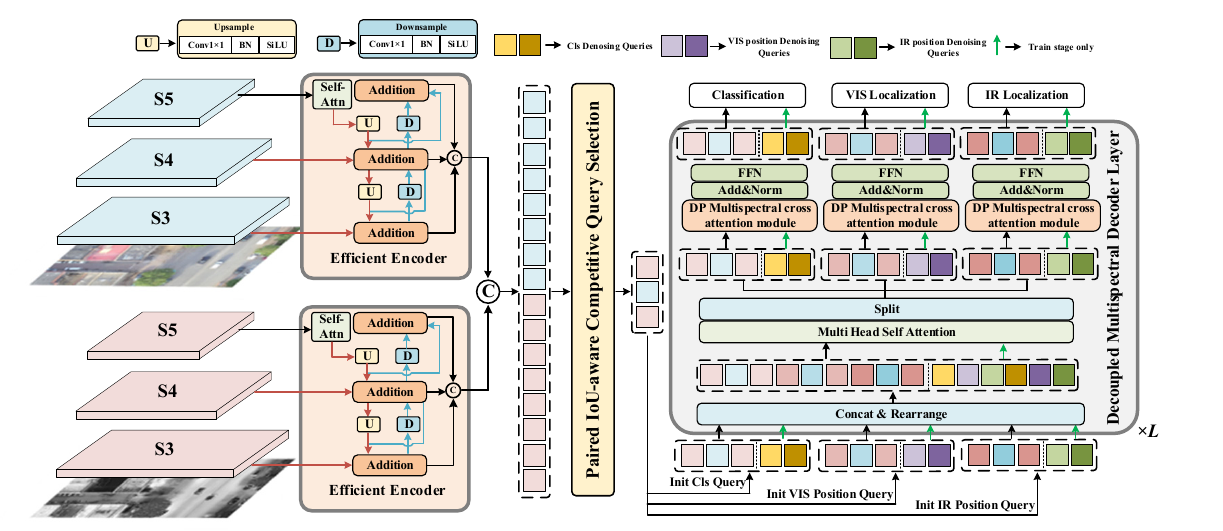}
    \caption{The overall architecture of  DPDETR. The framework consists of feature extraction and encoding modules, a Paired IoU-aware Competitive Query Selection module, a Decoupled Multispectral Decoder, and the Decoupled Position Contrastive DeNoising Training strategy. The final output of DPDETR is the category of each object and its reliable position in both visible and infrared images.}
    \label{fig:fig2_architecture}
\end{figure*}

\subsection{End to End Object Detectors}
In recent years, Carion et al. \cite{carion2020end} first proposed the end-to-end object detector based on transformer called DEtection TRansformer (DETR). It views object detection as a set prediction problem and uses binary matching to directly predict one-to-one object sets during training. However, DETR has the issue of slow training convergence and many DETR variants have been proposed to address this issue. Deformable DETR \cite{zhu2020deformable} accelerated training convergence by predicting 2D reference points and designing a Deformable cross-attention module to sparsely sample features around reference points. Conditional DETR \cite{meng2021conditional} decoupled the content and position information and proposed conditional cross-attention to accelerate training convergence. Efficient DETR \cite{yao2021efficient} built a more efficient pipeline by combining dense prediction and sparse prediction. DAB-DETR \cite{liu2022dab} introduced 4D reference points to optimize the anchor boxes layer by layer. DN-DETR \cite{li2022dn} accelerated the training process and label-matching effect by introducing query denoising training group. DINO \cite{zhangdino} integrated the above works to build a powerful DETR detection pipeline. Considering the computational efficiency of DETR, RT-DETR \cite{zhao2024detrs} achieves real-time object detection by designing an Efficient Hybrid Encoder.

Some works have also applied DETR to oriented object detection. O$^2$DETR \cite{ma2021oriented} was the first to apply DETR in this context, while AO$^2$-DETR \cite{dai2022ao2} introduced an oriented proposal refinement module for better adaptation to oriented object detection. ARS-DETR \cite{zeng2024ars} further proposed an angle-embedded Rotaed Deformable Attention module, which incorporates angle information for extracting the rotated object features in oriented object detection. 

Recently, a DETR-based infrared-visible object detection network called DAMSDet \cite{guo2024damsdet} was introduced, which addresses the complexity and variability of the complementary characteristics between infrared and visible modalities. 
In contrast, our DPDETR further addresses misalignment issues to achieve more reliable feature alignment.

\section{Proposed Method}
The overall architecture of the proposed DPDETR is shown in Fig. \ref{fig:fig2_architecture}. A pair of matched infrared and visible images are used as input.
The features of each image are first extracted by the modality-specific backbone network (e.g., ResNet50 \cite{he2016deep}), and then two modality-specific Efficient Encoders encode these features, respectively. Subsequently, the encoded features are flattened, concatenated, and input into a paired IoU-aware competitive query selection module. This module selects salient modality features that contain more accurate information on classification, visible modality position, and infrared modality position as the initial matching object queries. 
Next, we duplicate these matching object queries into three copies, which serve as the initial class query, initial visible position query, and initial infrared position query. These copies are then input into the Decoupled Multispectral Decoder for decoupled learning and feature alignment fusion to obtain refined decoupled matching queries. Finally, these refined matching queries are used to predict each object's classification, visible modality position, and infrared modality position.

Additionally, during the training stage, we implement the Decoupled Position DeNoising Training strategy to enhance the network’s decoupled learning ability and to increase the diversity of misalignment situations.
In the following subsections, we provide a detailed introduction to each module.

\begin{figure*}
    \centering
    \includegraphics[trim=22mm 58mm 22mm 62mm,clip,width=0.95\textwidth]{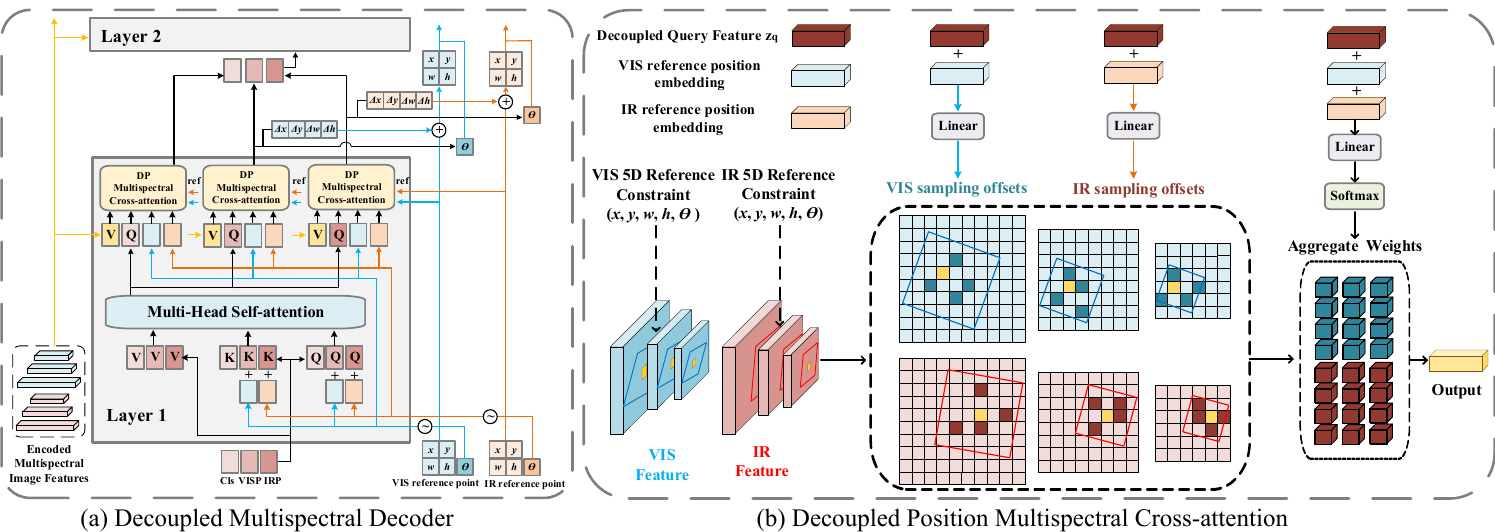}
    \caption{The structure of the Decoupled Multispectral Transformer Decoder (DeNoising Training Group is omitted in the figure) and Decoupled Position Multispectral Deformable cross-attention module.}
    \label{fig:fig3_decoder}
\end{figure*}

\subsection{Paired IoU-aware Competitive Query Selection}
The object queries in DETR \cite{carion2020end} are a set of learnable embeddings that contain the content and position information of the objects. In addition to setting object queries as learnable embeddings, several methods \cite{zhu2020deformable,yao2021efficient, zhangdino} use the confidence scores to select the Top-K features from the encoder to initialize object queries. The confidence score represents the likelihood that the feature includes foreground objects. 
In the infrared-visible object detection task, DAMSDet \cite{guo2024damsdet} achieves modality competitive query selection, comprehensively considering the classification and location confidence of feature representations to select modality-specific object queries.
However, we explicitly represent the object with category, visible modality position, and infrared modality position information. The detector is required to simultaneously model these three aspects of the objects, all of which determine the quality of the features. Hence, the confidence scores of the selected features should jointly reflect the object classification, visible modality position, and infrared modality position, rather than just representing the likelihood of foreground.
To implement this idea, we propose Paired IoU-aware Competitive Query Selection based on modality competitive query selection \cite{guo2024damsdet}. Concretely, we concatenate encoded feature sequences from the infrared and visible modalities and feed them into a linear projection layer to obtain classification scores. Then, we select the Top-$K$ scoring features as the initial object queries. This approach can be defined as follows:
\begin{equation}
    z=\operatorname{Top-\mathit{K}}(\operatorname{Linear}(\operatorname{concat}(I, V))),
\end{equation}
where $z$ denotes the set of $K$ selected modality-specific features, $I$ and $V$ represent the flattened encoded infrared and visible feature sequences, respectively. To ensure that the selected query $z$ simultaneously reflects a high classification score, visible modality position confidence, and infrared modality position confidence, we reformulate the optimization objective of the detector with the IoU of the two modality object positions as follows:

\vspace{-10pt}
\begin{small}
\begin{equation}
\begin{gathered}
L(\hat{y}, y)=L_{\text {boxvis }}\left(\hat{b}_{v}, b_{v}\right)+L_{\text {boxir }}\left(\hat{b}_{i}, b_{i}\right)+L_{c l s}\left(\hat{c}, \hat{b}_{v}, \hat{b}_{i}, c, b_{v}, b_{i}\right) \\
=L_{\text {boxvis }}\left(\hat{b}_{v}, b_{v}\right)+L_{b o x i r}\left(\hat{b}_{i}, b_{i}\right)+L_{c l s}\left(\hat{c}, c, I o U_{v i s}, I o U_{i r}\right),
\end{gathered}
\end{equation}
\end{small}
where $L_{\text {boxvis }}$,$L_{\text {boxir }}$, and $L_{\text {cls}}$ denote visible modality bounding boxes loss, infrared modality bounding boxes loss, and classification loss, respectively.
Here, $\hat{y}$ and $y$ denote prediction and ground truth, with $\hat{y}=\left\{\hat{c}, \hat{b}_{v}, \hat{b}_{i}\right\}$ and $y=\left\{c, b_{v}, b_{i}\right\}$. In this context, $c$, $b_{v}$, and $b_{i}$ represent categories, visible modality bounding boxes, and infrared modality bounding boxes, respectively. 
We introduce the visible and infrared IoU score into the loss function of the classification branch to realize the consistency constraint on the classification, visible modality localization, and infrared modality localization of selected queries.

\subsection{Decoupled Multispectral Transformer Decoder}

After the Paired IoU-aware competitive query selection, we obtain a set of initialization object queries that effectively represent the object category, visible modality position, and infrared modality position. However, there is a 
conflict in feature focus among
these three types of information. For example, category information tends to focus more on the central features of the object, while position information focuses more on the edge features \cite{zhang2023decoupled}. Additionally, in our infrared and visible modality scenarios, category information tends to incorporate features from both modalities, whereas modality-specific position information should focus more on features specific to each modality. This gap leads to limitations in the performance of our detectors when using the vanilla decoder structure, where a single set of queries simultaneously optimizes all three types of information and shares cross-attention. Hence, we design a decoupled decoder structure with decoupled queries and decoupled cross-attention branches for our optimization tasks to eliminate the gap between these three types of information.

\begin{figure}
    \centering
    \includegraphics[trim=110mm 89mm 110mm 89mm,clip,width=0.40\textwidth]{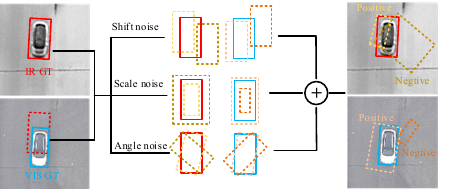}
    \caption{Demonstration of adding noise in terms of shift, scale, and angle to a pair of ground truth boxes. Red, light yellow, and dark yellow boxes represent the infrared GT box, positive infrared denoising box, and negative infrared denoising box, respectively. Blue, light orange, and dark orange boxes represent the visible GT box, positive visible denoising box, and negative visible denoising box, respectively.}
    \label{fig:fig4_add-noise}
\end{figure}

Specifically, the detailed structure of the Decoupled multispectral decoder is shown in Fig. \ref{fig:fig3_decoder} (a). Before inputting into the decoder, we obtain decoupled matching queries by duplicating the initial object queries into three sets: initial classification queries, visible modality position queries, and infrared modality position queries. In the decoupled multispectral decoder layer, we first concatenate the three types of matching queries together to apply self-attention, which can be formulated as follows:
\begin{small}
\begin{equation}
\left[\hat{q}_{c l s}, \hat{q}_{v i s p}, \hat{q}_{i r p}\right]=\operatorname{self-attn}\left(\operatorname{cat}\left(q_{c l s}, q_{v i s p}+E_{v i s p}, q_{i r p}+E_{i r p}\right)\right),
\end{equation}
\end{small}
where $q_{c l s}$, $q_{v i s p}$, and $q_{i r p}$ represent the classification queries, visible modality position queries, and infrared modality position queries, respectively. $E_{v i s p}$ and $E_{i r p}$ represent the object visible and infrared reference position embeddings, respectively. Through this self-attention approach, the network learns intra-class and global relationships across the three types of object information. Then, we feed the three types of self-attention queries, along with two modality object reference position embeddings, into three separate position decoupled cross-attention branches. Each branch searches for its matching interest modality and feature area, distilling relevant features and avoiding the impact of the optimization gap among the three types of information. 

Fig. \ref{fig:fig3_decoder} (b) shows the detailed structure of the Decoupled Position Multispectral Cross-attention. We add the query features with the visible modality reference position embeddings and use a linear layer to predict the sampling position offsets on the multi-scale visible feature maps. Similarly, we add the infrared modality reference position embeddings to predict the sampling position offsets on the multi-scale infrared feature maps. Then, we explicitly constrain the sampling range of both modalities using their respective reference positions to achieve aligned object feature sampling. Finally, we combine the query features with the two modality reference position embeddings to assign aggregation weights to the sampled feature points, aggregating them to obtain the output queries. Specifically, given the input multi-semantic infrared and visible feature maps $\left\{x_{v i s}^l, x_{i r}^l\right\}_{l=1}^L$, we use the $q$-th normalized centerpoint of visible and infrared reference positions $b_{q-vis}$ and $b_{q-ir}$ as the 2D visible reference point $\hat{p}_{q-v i s}$ and the infrared reference point $\hat{p}_{q-ir}$, respectively. We define the Position Decoupled Multispectral Deformable Cross-attention module $F$ as follows:
\begin{equation}
\begin{gathered}
F\left(z_q, \hat{p}_{q-v i s}, \hat{p}_{q-i r},\left\{x_{v i s}^l, x_{i r}^l\right\}_{l=1}^L\right)=\sum_h^H W_h \\
{\left[\sum_m \sum_{l=1}^L \sum_{k=1}^K A_{m h l q k} \cdot W_h^{\prime} x_m^l\left(\Phi_l\left(\hat{p}_{q-m}\right)+\Psi_l\left(\Delta \mathrm{p}_{m h l q k}\right)\right)\right]},
\end{gathered}
\end{equation}
where $z_{q}$ denotes the $q$-th query feature, $m \in\{v i s, i r\}$ denotes the visible and infrared modalities, $h$ indexes the attention head, $l$ indexes the input feature semantic level, and $k$ indexes the sampling point. $A_{mhlqk}$ and $\Delta \mathrm{p}_{m h l q k}$ denote the $k$-th attention weight and sampling point in the $l$-th feature semantic level and the $h$-th attention head within the $m$ modality, respectively. The attention weight $A_{mhlqk}$ is normalized by $\sum_m \sum_{l=1}^L \sum_{k=1}^K A_{m h l q k}=1$. The function $\Phi_l\left(\hat{p}_{q-m}\right)$ scales $\hat{p}_{q-v i s}$ and $\hat{p}_{q-i r}$ to the $l$-th semantic level feature map, while the function $\Psi_l\left(\Delta \mathrm{p}_{m h l q k}\right)$ constrains the predicted offset within the range of $b_{q-vis}$ and $b_{q-ir}$, thereby achieving object feature alignment.

\begin{figure}
    \centering
    \includegraphics[trim=92mm 78mm 92mm 75mm,clip,width=0.45\textwidth]{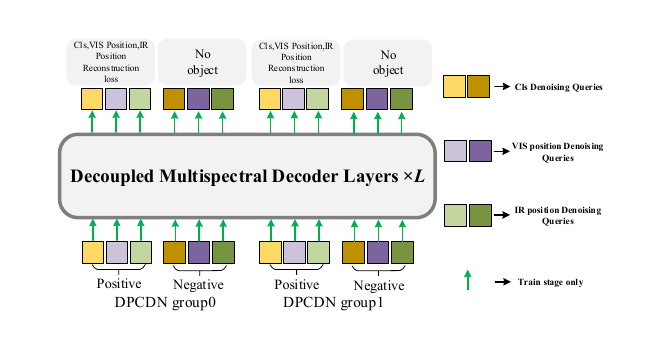}
    \caption{The structure of DPCDN group and the useage of positive and negative denoising queries. }
    \label{fig:fig5_DPCDN}
\end{figure}

For optimizing the object visible and infrared bounding boxes, we employ a cascaded optimization approach. Concretely, in the Decoupled Multispectral Decoder with D layers, we map from the $q$-th modality position query $z_{q-visp}^{d}$ and $z_{q-irp}^{d}$ in the $d$-th layer to obtain the refined infrared and visible reference bounding box $b_{q-vis}^d = [x,y,w,h]$ and $b_{q-ir}^d = [x,y,w,h]$. The process can be described as follows:
\begin{equation}
    b_{q-m\{x,y,w,h\}}^{d}=\sigma(MLP^{d}\Big(z_{q-m}^{d}\Big)+\sigma^{-1}(b_{q-m}^{d-1})),
\end{equation}
where $d\in\{2,3,...,D\}$, $MLP$ consists of two linear projection layers, $\sigma$ represents the sigmoid function, $\sigma^{-1}$ represents the inverse sigmoid function, and $b_{q-m}^1$ is the initialized bounding boxes from the Paired IoU-aware Competitive Query Selection stage. 
For the angles $\theta_{vis}$ and $\theta_{ir}$ of the bounding boxes in oriented object detection, we independently predict them by mapping visible position query features $z_{q-visp}^d$ and infrared position query features $z_{q-irp}^d$ at each layer.

\begin{figure}
    \centering
    \includegraphics[trim=78mm 35mm 73mm 35mm,clip,width=0.43\textwidth]{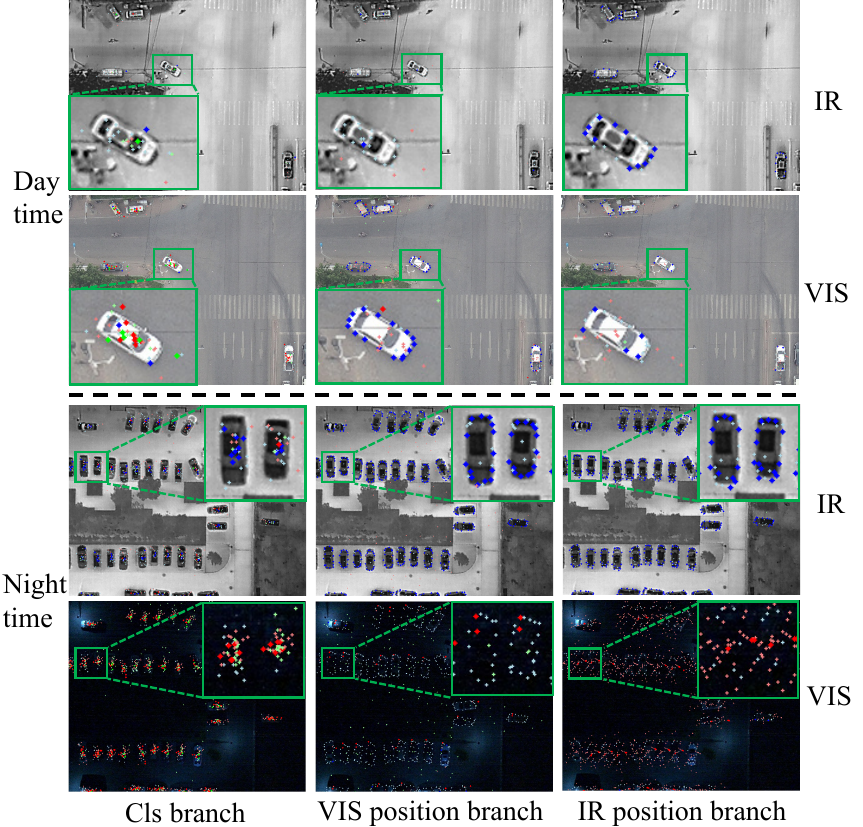}
    \caption{Visualization of infrared and visible feature points sampling of the three Decoupled Position Multispectral Cross-attention branches in the Decoupled Multispectral decoder. Different colors of points represent the results of sampling in different semantic layers, whereas blue, green, and red represent sampling points on low-level, middle-level, and high-level semantic feature maps, respectively. Dark colored and large points indicate relatively high aggregate weights.}
    \label{fig:fig6_decoupled_visul}
\end{figure}

\subsection{Decoupled Position Contrastive DeNoising Training}
In our network, simultaneously optimizing the three types of object information is more complex and largely depends on the accuracy of query-to-ground-truth matching. Therefore, inspired by DN-DETR \cite{li2022dn}, we design the Decoupled Position Contrastive DeNoising (DPCDN) training strategy. This approach bypasses the matching process and directly generates denoising queries by adding noise to the ground truth category, paired bounding box positions, sizes, and angles in both modalities. 

Specifically, we introduce category noise by randomly flipping ground truth labels to other labels, similar to DN-DETR \cite{li2022dn}. For paired visible and infrared bounding box noise, as illustrated in Fig. \ref{fig:fig4_add-noise}, we apply random shift, scale, and angle noise to the visible and infrared bounding boxes of an object to generate paired positive and negative two-modality denoising queries. The intensity of the noise is controlled by hyperparameters, with the noise for negative queries being larger than for positive queries. Random noise applied to the bounding boxes of both modalities also enhances the diversity of object misalignment situations between the two modalities.

As shown in Fig. \ref{fig:fig5_DPCDN}, we maintain consistency with the structure of the Decoupled Multispectral Decoder, decoupling positive and negative denoising queries into class, visible modality position, and infrared modality position denoising queries. If a pair of infrared and visible images has $n$ paired GT boxes, a DPCDN group will have 6 × $n$ queries with each paired GT box generating three decoupled positive queries and three decoupled negative queries. We also use multiple DPCDN groups to improve the effectiveness of our method. These denoising queries undergo self-attention and cross-attention operations in the decoder with matching queries. To prevent information leakage during self-attention, we introduce the attention mask \cite{li2022dn} to ensure that matching queries cannot see denoising queries, and denoising groups cannot see each other. Finally, these positive queries are responsible for reconstructing the corresponding object GT category, visible modality, and infrared modality boxes, while negative queries are expected to predict the background. The reconstruction losses are L1 and PIOU (GIOU for horizontal object detection) losses for box regression and focal loss \cite{lin2017focal} for classification. 

\begin{figure}
    \centering
    \includegraphics[trim=75mm 66mm 89mm 73mm,clip,width=0.45\textwidth]{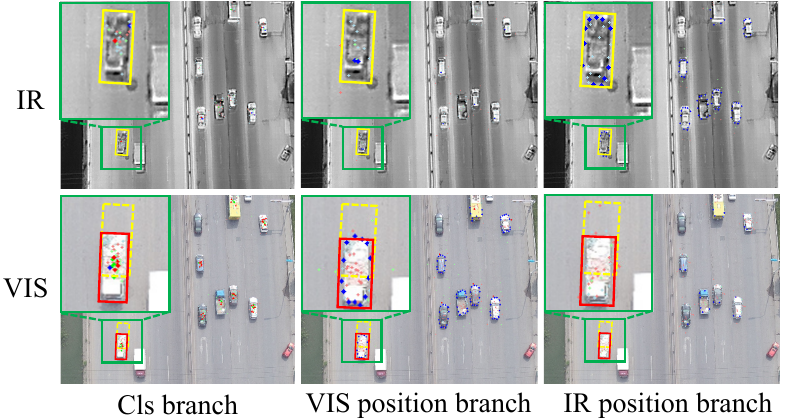}
    \caption{Visualization of infrared and visible feature points sampling of the three Decoupled Position Multispectral Cross-attention branches in the misalignment scene. The yellow and red boxes represent the position of the same object in the infrared and visible modalities, respectively.}
    \label{fig:fig7_misalignment_visul}
\end{figure}

\subsection{Matching Cost and Loss Function}
Consistent with the DETR-like approach, we calculate a bipartite matching between the ground truth set of objects $y$ and the set of $N$ predictions $\hat{y}$ for a permutation of $N$ elements $\sigma\in\mathfrak{S}_{N}$ with the lowest cost:
\begin{equation}
    \hat{\sigma}=\underset{\sigma\in\mathfrak{S}_N}{\arg\min}\sum_i^N\mathcal{L}_{\mathrm{match}}(y_i,\hat{y}_{\sigma(i)}),
\end{equation}
where $\mathfrak{S}_{N}$ represents the set of all possible permutations, $\mathcal{L}_{\mathrm{match}}(y_i,\hat{y}_{\sigma(i)})$ is the matching cost between ground truth $y_{i}$ and a prediction with index $\sigma(i)$. This process is computed with the Hungarian algorithm. Since we explicitly optimize the object category, visible modality position and infrared modality position, we define $y_{i}=(c_{i}, b_{vis-i}, b_{ir-i})$ where $c_{i}$ is the object class label, $b_{vis-i}$ and $b_{ir-i}$ are the 5D vectors that represent visible and infrared ground truth oriented box, respectively. Thus we define $\mathcal{L}_{\mathrm{match}}(y_i,\hat{y}_{\sigma(i)})$ as $-\mathds{1}_{\{c_{i}\neq\emptyset\}}\hat{p}_{\sigma(i)}(c_{i})+\mathds{1}_{\{c_{i}\neq\emptyset\}}\mathcal{L}_{\mathrm{box}}\big(b_{\nu is-i},\hat{b}_{\nu is-\sigma(i)}\big)+\mathds{1}_{\{c_{i}\neq\emptyset\}}\mathcal{L}_{\mathrm{box}}(b_{ir-i},\hat{b}_{ir-\sigma(i)})$ to ensure that the matched queries can reliably locate the objects in both the visible and infrared modalities. Where $L_{box}$ is composed of L1 cost and PIoU \cite{chen2020piou} cost for oriented box matching cost calculation.

\begin{table*}[]
    \setlength{\tabcolsep}{12pt}
    \renewcommand{\arraystretch}{1.1}
    \centering
\caption{Detection results ($m$AP, in \%) on DroneVehicle Dataset. The best results are highlighted in \textbf{bold}. }
\label{tab:tab4_dronevechal_result}
    \begin{tabular}{lcccccccl} 
         \toprule
         Model & 
      Car& Truck& Bus&Van &Freight-car&$m$AP &Speed(fps)&Modality\\
       \midrule
 Faster R-CNN (OBB) \cite{ren2016faster} & 79.69& 41.99& 76.94&37.68& 33.99&54.06 &&\multirow{6}*{VIS}\\
 RetinaNet (OBB) \cite{lin2017focal}& 78.45& 34.39& 69.75&28.82& 24.14&47.11 &-&\\
 S$^2$ANet \cite{han2021align} & 79.86& 50.02& 82.77&37.52& 36.21&57.28 &-&\\
 Oriented R-CNN \cite{xie2021oriented}& 80.26& 55.39& 86.84&46.92& 42.12&62.30 &-&\\
 ROI Transformer \cite{ding2019learning}& 61.55& 55.05& 85.48&44.84& 42.26&61.55 &-&\\
 Oriented RepPoints \cite{li2022oriented}& 84.40& 55.00& 85.80&46.60& 39.50&62.30 &-&\\
 \midrule
 Faster R-CNN (OBB) \cite{ren2016faster}& 89.68& 40.95& 86.32&41.21& 43.10&60.27 &-&\multirow{6}*{IR}\\
 RetinaNet (OBB) \cite{lin2017focal}& 88.81& 35.43& 76.45&32.12& 39.47&54.45 &-&\\
 
S$^2$ANet \cite{han2021align}& 89.71& 51.03& 88.97& 44.03& 50.27& 64.80 &-&\\
 
Oriented R-CNN \cite{xie2021oriented}& 89.63& 53.92& 89.15& 40.95& 53.86& 65.50 &-&\\
 
ROI Transformer \cite{ding2019learning}& 89.65& 50.98& 88.86& 44.47& 53.42& 65.47 &-&\\
 
Oriented RepPoints \cite{li2022oriented}& 89.90& 55.60& 89.10& 48.10& 57.60& 68.00 &-&\\ 
 \midrule
 Halfway Fusion (OBB) \cite{wagner2016multispectral}& 89.85& 60.34& 88.97& 46.28& 55.51& 68.19 &20.4&\multirow{7}*{VIS+IR}\\
 CIAN (OBB) \cite{zhang2019cross}& 89.98& 62.47& 88.90& 49.59& 60.22& 70.23 &21.7&\\
 AR-CNN (OBB) \cite{zhang2021weakly}& 90.08& 64.82& 89.38& 51.51& 62.12& 71.58 &18.2&\\
 TSFADet \cite{yuan2022translation}& 89.88& 67.87& 89.81& 53.99& 63.74& 73.06 &18.6&\\
 C$^2$Former \cite{yuan2024c}& 90.20& 68.30& 89.80& 58.50& 64.40& 74.20 &-&\\
 GAGTDet \cite{yuan2024improving}& \textbf{90.82}& 69.65& \textbf{90.46}& 55.62& 66.28& 74.57 &17.8&\\
 DAMSDet (OBB) \cite{guo2024damsdet}& 90.30& 74.90& 89.70& 63.20& 70.70& 77.75& 11.4&\\
 DPDETR (Ours)& 90.30& \textbf{78.20}& 90.10& \textbf{64.90}& \textbf{75.70}& \textbf{79.81} &7.2&\\
 \bottomrule
 \end{tabular}
\end{table*}

For the total loss function of DPDETR, we define it as follows:
\begin{equation}
    \mathcal{L}=\mathcal{L}_{cls}+\mathcal{L}_{box-vis}+\mathcal{L}_{box-ir}+\mathcal{L}_{dn}+\mathcal{L}_{Aux},
\end{equation}
where $L_{cls}$ is the Paired IoU-aware classification loss, $L_{box-vis}$ and $L_{box-ir}$ are composed of L1 loss and PIoU loss for visible and infrared oriented box regression, $L_{dn}$ is Decoupled Position Contrastive DeNoising Training loss, and $L_{Aux}$ is the auxiliary loss computed at the output of each decoder layer.

\begin{table}
    \centering
\caption{Ablation experiments for Decoupled Position Constructive DeNoising training and Query Decoupled Structure on DroneVehicle Dataset. }
\label{tab:tab1_ablation_study}
    \begin{tabular}{cccccc}
         \toprule  
        &  Model&DPCDN& QDS& $m$AP&Modality\\
  \midrule
   1&  Our Base Detector&-& -& 70.40&VIS\\
  2&  Our Base Detector&-& -& 76.06&IR\\
 \midrule
   3&  DPDETR&-& -& 77.40&\multirow{4}*{VIS+IR}\\
   4&  DPDETR&\checkmark& -& 78.03&\\
   5&  DPDETR&-& \checkmark& 78.92&\\
   6&  DPDETR&\checkmark& \checkmark& \textbf{79.81}&\\
  \bottomrule
 \end{tabular}
\end{table}

\section{Experiments}
In this section, we first introduce the datasets and evaluation metrics used in our experiments. We then provide comparisons with several state-of-the-art methods. Next, we conduct ablation studies and perform a detailed analysis of the effectiveness of our methods. Finally, we analyze and compare the computational cost of our method.

\subsection{Datasets and Evaluation Metric}
We perform experiments using two publicly available datasets for infrared-visible object detection:  the DroneVehicle Dataset \cite{sun2022drone} and the KAIST Multispectral Pedestrian Detection Dataset \cite{hwang2015multispectral}. Both datasets include pairs of VIS-IR images and annotations for conducting experiments.

\textbf{DroneVehicle dataset.} The DroneVehicle dataset is a large-scale drone-based infrared-visible oriented object detection dataset. It contains 28,439 pairs of VIS-IR images, covering various scenarios such as urban roads, residential areas, parking lots, and different times from day to night. The dataset provides independently annotated oriented bounding box labels for infrared and visible modalities, covering five categories: car, truck, bus, freight car, and van. To validate the efficacy of our method, we refer to \cite{yuan2024improving} and make some adjustments to the original infrared and visible annotations:
\begin{itemize}
\item Objects annotated in only one modality will be inserted into the same positions in the other modality.
\item Objects in the infrared and visible modalities are organized to ensure consistent indexing. 
\end{itemize}
The adjusted dataset includes 17,990 image pairs for training and 1,469 pairs for testing.



\begin{table}[]
    \centering
    \setlength{\tabcolsep}{12pt}
\caption{Ablation experiments for query initialization strategies on DroneVehicle Dataset. }
\label{tab:tab2_ablation_query_initial}
    \begin{tabular}{ccc}
         \toprule  
          Learnable Queries& PICQS& $m$AP\\
  \midrule
     \checkmark& -& 77.48\\
    -& \checkmark& \textbf{79.81}\\
  \bottomrule
 \end{tabular}
\end{table}

\textbf{KAIST dataset.} The KAIST dataset is a large-scale infrared-visible pedestrian detection dataset. It contains 95,328 pairs of images captured in driving environments with traffic scenes including streets, campuses, and countryside. Since the original dataset had annotation issues, we use new annotations provided by \cite{zhang2019weakly} and inferred using the improved annotations \cite{liu2016multispectral}. The training set consists of 7,601 pairs of images, while the test set includes 2,252 pairs, among which 1,455 pairs are from daytime scenes and 797 pairs are from nighttime scenes.

\textbf{Evaluation metric.} For the DroneVehicle Dataset, we evaluate detection performance using mean average precision ($m$AP). Specifically, we calculate $m$AP using True Positive (TP) and False Positive (FP) at the Intersection over Union (IoU) threshold of 0.5. For the KAIST Dataset, we adopt log average miss rate $MR^{-2}$ over the false positive per image (FPPI) with a range of $[10^{-2},10^0]$ to evaluate the pedestrian detection performance. A lower value is better. To further evaluate the performance of the method, we assess its effectiveness under the `All' condition \cite{hwang2015multispectral} and across six all-day subsets from the test set, which include variations in pedestrian distances and occlusion levels.

\begin{table*}[]
    \setlength{\tabcolsep}{12pt}
    \renewcommand{\arraystretch}{1.1}
    \centering
\caption{Detection results (MR, in\%) under the IoU threshold (0.5) of different pedestrian distances, occlusion levels and light conditions (Day and Night) on KAIST Dataset. The pedestrian distances consist of `Near' (115 $\leq$ \textit{height}), `Medium' (45 $\leq$ \textit{height} \textless 115) and `Far' (1 $\leq$ \textit{height} \textless 45), with occlusion levels consist of `None' (never occluded), `Partial' (occluded to some extent up to one half) and `Heavy' (mostly occluded). The best results are highlighted in \textbf{bold}. A lower value is better.}
\label{tab:tab6_kaist_result}
    \begin{tabular}{lccccccccc} 
    \toprule
    Model & Near & Medium & Far & None & Partial & Heavy & Day & Night & All \\
    \midrule
    ACF \cite{hwang2015multispectral}& 28.74& 53.67& 88.20& 62.94& 81.40& 88.08& 64.31& 75.06& 67.74\\
    Halfway Fusion \cite{wagner2016multispectral}& 8.13& 30.34& 75.70& 43.13& 65.21& 74.36& 47.58& 52.35& 49.18\\
    FusionRPN+BF \cite{konig2017fully}& 0.04& 30.87& 88.86& 47.45& 56.10& 72.20& 52.33& 51.09& 51.70\\
    IAF R-CNN \cite{li2019illumination}& 0.96& 25.54& 77.84& 40.17& 48.40& 69.76& 42.46& 47.70& 44.23\\
    IATDNN+IASS \cite{guan2019fusion}& 0.04& 28.55& 83.42& 45.43& 46.25& 64.57& 49.02& 49.37& 48.96\\
    CIAN \cite{zhang2019cross}& 3.71& 19.04& 55.82& 30.31& 41.57& 62.48& 36.02& 32.38& 35.53\\
    MSDS-RCNN \cite{li2018multispectral}& 1.29& 16.19& 63.73& 29.86& 38.71& 63.37& 32.06& 38.83& 34.15\\
    AR-CNN \cite{zhang2019weakly}& 0.00& 16.08& 69.00& 31.40& 38.63& 55.73& 34.36& 36.12& 34.95\\
    MBNet \cite{zhou2020improving}& 0.00& 16.07& 55.99& 27.74& 35.43& 59.14& 32.37& 30.95& 31.87\\
    TSFADet \cite{yuan2022translation}& 0.00& 15.99& 50.71& 25.63& 37.29& 65.67& 31.76& 27.44& 30.74\\
 DAMSDet\cite{guo2024damsdet}& 1.47& 20.88& 42.62& 26.28& 34.45& 58.45& 32.84& 22.21&30.27\\
    CMPD \cite{li2022confidence}& 0.00& \textbf{12.99}& 51.22& 24.04& 33.88& 59.37& 28.30& 30.56& 28.98\\
    GAGTDet \cite{yuan2024improving}& 0.00& 14.00& 49.40& 24.48& 33.20& 59.35& 28.79& 27.73& 28.96\\
    C$^2$Former \cite{yuan2024c}& 0.00& 13.71& 48.14& 23.91& 32.84& 57.81& 28.48& 26.67& 28.39\\
    \midrule
    DPDETR (Ours)& \textbf{0.00}& 15.54& \textbf{36.84}& \textbf{21.09}& \textbf{26.61}& \textbf{57.05}& \textbf{26.75}& \textbf{21.01}& \textbf{25.04}\\
    \bottomrule
\end{tabular}
\end{table*}

\begin{table}[]
    \centering
    \setlength{\tabcolsep}{12pt}
\caption{Ablation experiments for the number of decoder layers on DroneVehicle Dataset. }
\label{tab:tab3_ablation_decoder_layers}
    \begin{tabular}{ccc}
         \toprule  
          Decoder layers num& $m$AP &Params\\
  \midrule
     4& 77.38&84.1M\\
  5& 78.47&87.1M\\
  6& \textbf{79.81}&\textbf{90.1M}\\
    7& 78.94&93.1M\\
  \bottomrule
 \end{tabular}
\end{table}

\subsection{Implementation Details}
For the DroneVehicle Dataset, we employ our oriented object detection version, while for the KAIST Dataset, we remove the angle prediction components to achieve horizontal object detection. We select ResNet50 \cite{he2016deep} as the backbone of both the infrared and visible branches, with a feature map semantic level of $L$=3. The Decoupled Multispectral Decoder contains six layers and we set the number of attention heads, sampling points, and selected queries as $H$=8, $K$=4, and $N$=300, respectively. During training, we employ basic data augmentations, including random rotation, and flipping, with both training and testing image sizes set to 640 × 640. For the DroneVehicle dataset, we load ResNet50 pre-trained weights and train it for 42 epochs. For the KAIST dataset, we load our base detector pre-trained on the COCO dataset and train it for 24 epochs. The optimizer used is Adam, configured with a learning rate of 0.0001 and a weight decay of 0.0001. All experiments are conducted on an NVIDIA A100 GPU.  


\subsection{Evaluation on DroneVehicle Dataset}
As for infrared and visible oriented object detection, we compare our method with six state-of-the-art single-modality detectors, including Faster R-CNN \cite{ren2016faster}, RetinaNet \cite{lin2017focal}, S$^2$ANet \cite{han2021align}, Oriented R-CNN \cite{xie2021oriented}, ROI Transformer \cite{ding2019learning}, and Oriented RepPoints \cite{li2022oriented}. We also make a comparison with 7 infrared-visible multispectral object detection methods, including Halfway Fusion \cite{liu2016multispectral}, CIAN \cite{zhang2019cross}, AR-CNN \cite{zhang2019weakly}, TSFADet \cite{yuan2022translation}, C2Former \cite{yuan2024c}, GAGTDet \cite{yuan2024improving}, and DAMSDet \cite{guo2024damsdet}. Among them, Halfway Fusion (OBB) is a multispectral oriented object detection method constructed by adding the features of a dual-branch S2ANet. AR-CNN \cite{zhang2021weakly}, TSFADet \cite{yuan2022translation}, and GAGTDet \cite{yuan2024improving} all use both infrared and visible object annotations for constraint. Table \ref{tab:tab4_dronevechal_result} shows the comparison results of these methods. For single-modality detection results, the infrared modality significantly outperforms the visible modality because objects in nighttime scenes are difficult to distinguish from the visible modality. 
Obviously, infrared-visible multispectral object detection methods are superior to single-modality methods.  Our method approaches the SOTA method in the `Car' and `Bus' categories, where the metrics are relatively saturated. It shows significant improvements in the `Truck,' `Van,' and `Freight-car' categories, which are more difficult to distinguish, ultimately surpassing the best method, DAMSDet, by 2.06 $m$AP. 

\begin{figure*}
    \centering
    \includegraphics[trim=55mm 10mm 45mm 7mm,clip,width=0.9\textwidth]{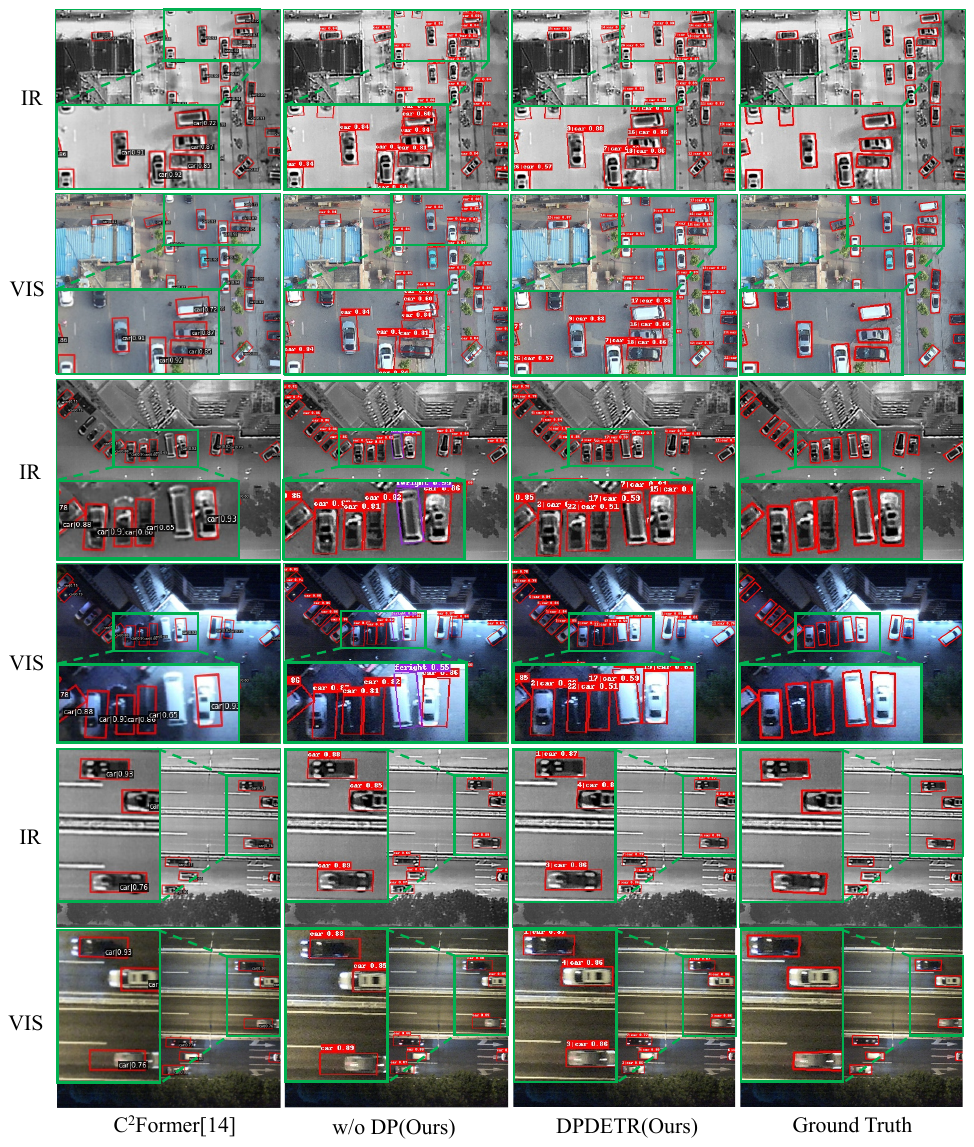}
    \caption{Detection Results on the DroneVehicle Dataset. The w/o DP method represents without decoupled position and only uses infrared modality annotations to constrain. The confidence threshold is set to 0.5 when visualizing these results. }
    \label{fig:fig8_dronechicle_detection}
\end{figure*}

We also provide some detection visualization results on the DroneVehicle Datasets. As shown in Figure \ref{fig:fig8_dronechicle_detection}, C$^2$Former \cite{yuan2024c} and w/o DP can only output the locations of objects in one modality under misalignment conditions, failing to locate objects in the other modality accurately. The w/o DP even encounters duplicate detection problems due to obvious misalignment situations. In contrast, our DPDETR can reliably locate objects in both modalities and establish the correspondence of the same object in complex misalignment scenarios caused by imaging and motion displacement. DPDETR also achieves better classification and localization by fully leveraging the misaligned object features.

\subsection{Ablation Studies}
We conduct detailed ablation experiments on the DroneVehicle dataset to demonstrate the effectiveness of our method. 

\textbf{Ablations on Decoupled Position Contrastive DeNoising (DPCDN) training and Query Decoupled Structure (QDS).} In Table \ref{tab:tab1_ablation_study}, the 1\textsuperscript{st} and 2\textsuperscript{nd} rows display the detection performance of our single modality base detector for infrared and visible modalities, respectively. The 3\textsuperscript{rd} row shows the performance of our base DPDETR on the VIS-IR dual modality. Our base DPDETR outperforms each single modality performance, but it only shows a 1.34 $m$AP improvement over the IR modality due to the challenges in optimizing the decoupled position and correspondence between the two modalities. As shown in the 4\textsuperscript{th} and 5\textsuperscript{th} rows of Table \ref{tab:tab1_ablation_study}, we add DPCDN and DQS to achieve an additional improvement of 0.63 $m$AP and 1.52 $m$AP, respectively. Finally, by combining DPCDN and DQS, our method reaches 79.81 $m$AP, representing a 3.75 $m$AP improvement over the IR modality.  

\begin{figure}
    \centering
    \includegraphics[trim=55mm 38mm 55mm 32mm,clip,width=0.475\textwidth]{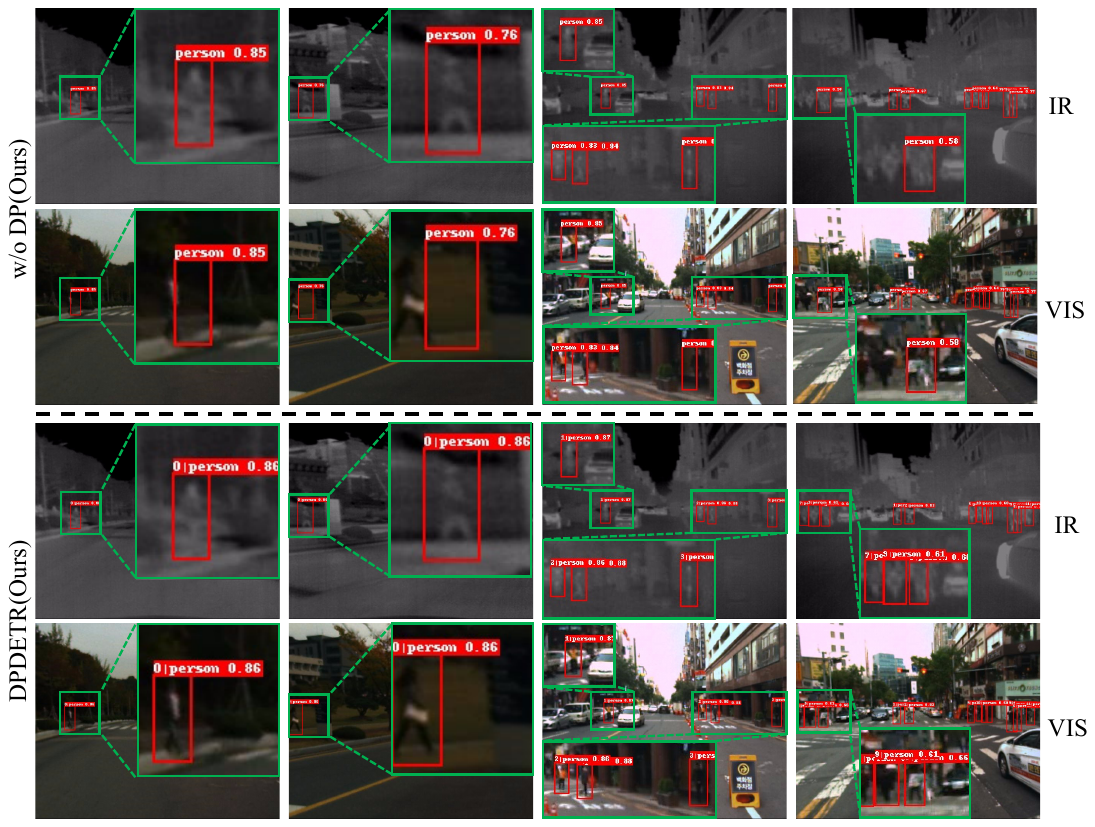}
    \caption{Detection Results on the KAIST Dataset. The confidence threshold is set to 0.5 when visualizing these results.}
    \label{fig:fig9_kaist_detection}
\end{figure}

\textbf{Ablations on query initialization.} We conduct ablation experiments on query initialization strategies. Specifically, we set three types of decoupled queries as three groups of learnable embeddings for comparison with our Paired IoU-aware Competitive Query Selection (PICQS) initialization method.
As shown in Table \ref{tab:tab2_ablation_query_initial}, our PICQS method outperforms the learnable queries by 2.33 $m$AP. We attribute this to the complex and varied complementary characteristics of infrared and visible images, thus making our approach of competitively selecting object queries based on image pair characteristics more effective than using pre-learned fixed object queries.



\textbf{Ablations on the number of decoder layers.} We also conduct ablation experiments on the number of Decoupled Multispectral Decoder layers. As shown in Table \ref{tab:tab3_ablation_decoder_layers}, we find that the best results are achieved when the number of layers is set to 6.

\subsection{Effectiveness Analysis of Our Method}
To illustrate the effectiveness of our method, we first visualize the adaptive infrared and visible feature point sampling of the three Decoupled Position Multispectral Cross-attention branches in the Decoupled Multispectral decoder for both daytime and nighttime scenes in Figure \ref{fig:fig6_decoupled_visul}. 
In daytime scenarios, the classification branch focuses more on the high-level semantic internal features of the object, primarily attending to the high-level semantic features of the visible modality. This is because visible features contain richer abstract high-level semantic information, which is crucial for classification. In contrast, both the visible and infrared position branches focus more on the low-level semantic contour features of the object, which are more important for object localization. Specifically, the visible position branch mainly focuses on the visible features, while the infrared position branch primarily focuses on the infrared features. 
In nighttime scenes, the classification branch adaptively increases its focus on the infrared modality features to ensure correct classification since the object features in the visible modality are very weak. In the Visible Position branch, the network optimizes the visible position by focusing on the contour features of the infrared modality objects, reflecting the effectiveness of our method in adaptive complementary feature fusion. In the infrared Position branch, the network also correctly focuses on the object contour features of the infrared modality.

Additionally, in Figure \ref{fig:fig7_misalignment_visul}, we visualize the effectiveness of our Decoupled Position Multispectral Cross-attention in achieving aligned feature fusion under a misalignment scene. In all three decoupled branches, our network correctly focuses on the correct positions of the same object in both modalities.

\subsection{Evaluation on KAIST Dataset}
As for infrared and visible pedestrian detection, we remove the angle prediction components of our DPDETR to achieve horizontal object detection. In the experiments, we compare our method with several state-of-the-art multispectral pedestrian detection methods, including ACF \cite{wagner2016multispectral}, Halfway Fusion \cite{liu2016multispectral}, FusionRPN+BF \cite{konig2017fully}, IAF R-CNN \cite{li2019illumination}, IATDNN+IASS \cite{guan2019fusion}, CIAN \cite{zhang2019cross}, MSDS-RCNN \cite{li2018multispectral}, AR-CNN \cite{zhang2019weakly}, MBNet \cite{zhou2020improving}, TSFADet \cite{yuan2022translation}, CMPD \cite{li2022confidence}, GAGTDet \cite{yuan2024improving}, DAMSDet\cite{guo2024damsdet}, and C$^2$Former \cite{yuan2024c}. As shown in Table \ref{tab:tab6_kaist_result}, our method obtains 26.75 MR, 21.01 MR, and 25.04 MR on the `Day,' `Night,' and `All' conditions under the IoU threshold of 0.5, respectively. Specifically, Our method outperforms the previous best method C$^2$Former \cite{yuan2024c} by 1.73 MR, 5.66 MR, and 3.35 MR in the `Day,' `Night,' and `All' conditions. Especially, the significant 5.66MR improvement under `Night' conditions indicates that our method is effective in adaptively fusing complementary features and avoiding interference from visible modality features at nighttime.

\begin{table}[]
    \centering
\caption{Computational cost comparison for DPDETR with SOTA methods. }
\label{tab:tab5_computational_cost}
    \begin{tabular}{ccccc}
         \toprule  
          Methods&$m$AP& FLOPs&Params &Speed(fps)\\
  \midrule
     TSFADet \cite{yuan2022translation}&73.06& 109.8G&104.7M &18.6\\
 GAGTDet \cite{yuan2024improving}& 74.57& 120.6G&- &17.8\\
 C$^2$Former \cite{yuan2024c}& 74.20& 177.7G&120.8M &-\\
 DAMSDet (OBB)\cite{guo2024damsdet}& 78.24& 182.7G&77.5M &11.4\\
    DPDETR (Ours)&\textbf{79.81}& \textbf{208.0G}&\textbf{90.1M} &\textbf{7.2}\\
  \bottomrule
 \end{tabular}
\end{table}

We also provide some detection visualization results on the KAIST Dataset. As shown in Figure \ref{fig:fig9_kaist_detection}, our DPDETR can reliably locate the pedestrians in both modalities and achieve higher detection confidence even in the case of obvious misalignment scenarios.

\begin{figure}
    \centering
    \includegraphics[trim=30mm 58mm 20mm 50mm,clip,width=0.475\textwidth]{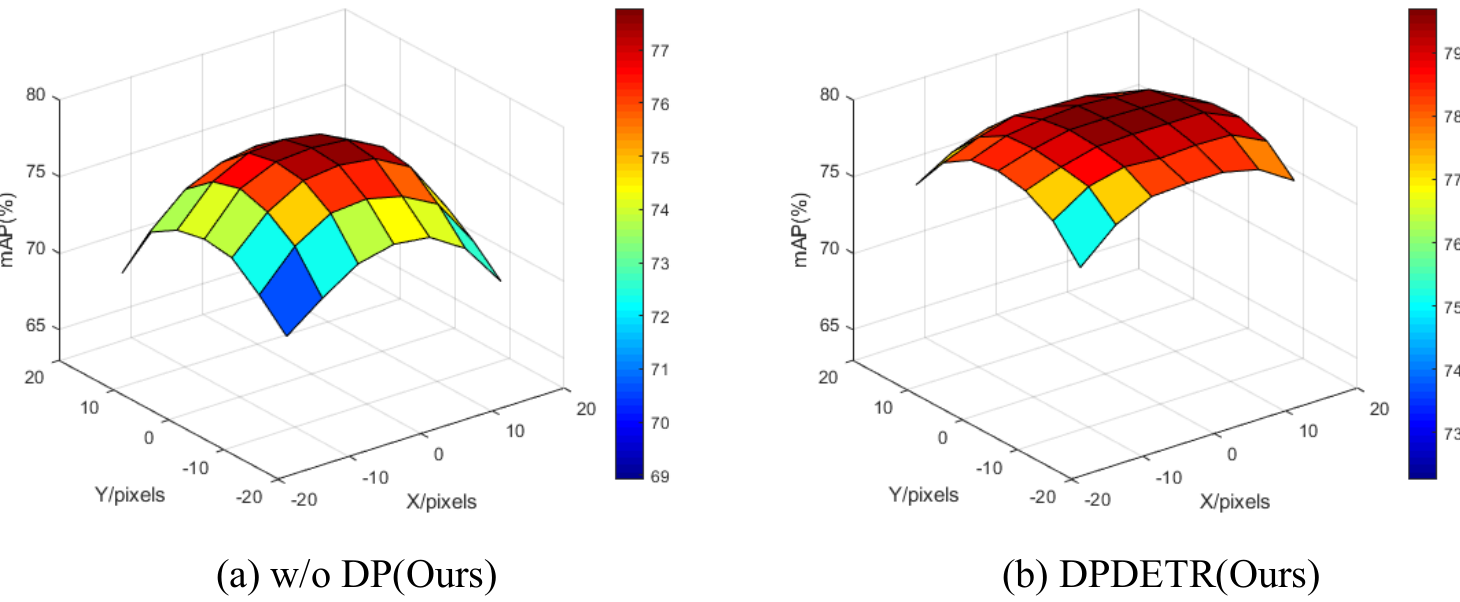}
    \caption{Visualization of the manually misalignment experiments. The position shift step size along both the x-axis and y-axis is 5 pixels.}
    \label{fig:fig10_shiftaug}
\end{figure}

\subsection{Robustness to the more Serious Misalignment Scenarios}
To further evaluate the robustness of the proposed method under more serious modality misalignment problem, we conduct experiments on the DroneVehicle dataset by manually simulating more serious misalignment scenarios. Specifically, we perform both training and evaluation on w/o DP method and DPDETR. In the training phase, we introduce synthetic more serious misalignment by applying position shifts to paired images. Concretely, for each image pair, we randomly select either the infrared or visible image as the Position Shift Image (PSI). The PSI is then shifted along the x-axis by a random offset sampled from the range $[-13,13]$ pixels. Additionally, with a certain probability, a similar shift is applied along the y-axis to further simulate real-world misalignment conditions. During evaluation, we fix the infrared image and apply a position shift to the visible image by $[\Delta x, \Delta y]$ pixels, where $\Delta x, \Delta y \in [-15, 15]$. The experimental results are illustrated in Figure \ref{fig:fig10_shiftaug}. It can be seen that our method achieves a mAP of 79.70 under simulated more serious modality misalignment, and the performance degradation of DPDETR on the 
more serious misaligned validation situations remains minimal. These results demonstrate that our approach exhibits strong robustness to modality misalignment during both training and inference.

\subsection{Computational Cost Comparison}
As shown in Table \ref{tab:tab5_computational_cost}, compared to other SOTA multispectral object detection methods, our DPDETR does not significantly increase the computational cost (GFLOPs) and even has fewer parameters than these methods. However, since DPDETR simultaneously outputs the locations of objects in both infrared and visible modalities, the inference speed is slightly slower.

\section{Conclusion}
In this paper, we propose DPDETR to address the modality misalignment problem by explicitly decoupling and optimizing the object's position and correspondence in infrared-visible object detection. Through the query decoupled structure, DPDETR effectively focuses on different multispectral complementary features for the object category, visible position, and infrared position, eliminating the optimization gap. In the Position Decoupled Multispectral Deformable Cross-attention module, we achieve reliable adaptive multi-sematic fusion of misaligned features with the constraint of objects' visible and infrared reference positions. Additionally, our Decoupled Position Contrastive DeNoising Training further enhances DPDETR's decoupled learning capability. Experiments on drone-vehicle detection and pedestrian detection demonstrate that the proposed method achieves significant improvements compared to other state-of-the-art methods.



\bibliographystyle{IEEEtran}
\bibliography{main.bib}

\begin{thebibliography}{10}
\providecommand{\url}[1]{#1}
\csname url@samestyle\endcsname
\providecommand{\newblock}{\relax}
\providecommand{\bibinfo}[2]{#2}
\providecommand{\BIBentrySTDinterwordspacing}{\spaceskip=0pt\relax}
\providecommand{\BIBentryALTinterwordstretchfactor}{4}
\providecommand{\BIBentryALTinterwordspacing}{\spaceskip=\fontdimen2\font plus
\BIBentryALTinterwordstretchfactor\fontdimen3\font minus \fontdimen4\font\relax}
\providecommand{\BIBforeignlanguage}[2]{{%
\expandafter\ifx\csname l@#1\endcsname\relax
\typeout{** WARNING: IEEEtran.bst: No hyphenation pattern has been}%
\typeout{** loaded for the language `#1'. Using the pattern for}%
\typeout{** the default language instead.}%
\else
\language=\csname l@#1\endcsname
\fi
#2}}
\providecommand{\BIBdecl}{\relax}
\BIBdecl

\bibitem{ren2016faster}
S.~Ren, K.~He, R.~Girshick, and J.~Sun, ``Faster r-cnn: Towards real-time object detection with region proposal networks,'' \emph{IEEE transactions on pattern analysis and machine intelligence}, vol.~39, no.~6, pp. 1137--1149, 2016.

\bibitem{liu2016ssd}
W.~Liu, D.~Anguelov, D.~Erhan, C.~Szegedy, S.~Reed, C.-Y. Fu, and A.~C. Berg, ``Ssd: Single shot multibox detector,'' in \emph{Computer Vision--ECCV 2016: 14th European Conference, Amsterdam, The Netherlands, October 11--14, 2016, Proceedings, Part I 14}.\hskip 1em plus 0.5em minus 0.4em\relax Springer, 2016, pp. 21--37.

\bibitem{redmon2018yolov3}
J.~Redmon and A.~Farhadi, ``Yolov3: An incremental improvement,'' \emph{arXiv preprint arXiv:1804.02767}, 2018.

\bibitem{herrmann2018cnn}
C.~Herrmann, M.~Ruf, and J.~Beyerer, ``Cnn-based thermal infrared person detection by domain adaptation,'' in \emph{Autonomous Systems: Sensors, Vehicles, Security, and the Internet of Everything}, vol. 10643.\hskip 1em plus 0.5em minus 0.4em\relax SPIE, 2018, pp. 38--43.

\bibitem{kieu2020task}
M.~Kieu, A.~D. Bagdanov, M.~Bertini, and A.~Del~Bimbo, ``Task-conditioned domain adaptation for pedestrian detection in thermal imagery,'' in \emph{European conference on computer vision}.\hskip 1em plus 0.5em minus 0.4em\relax Springer, 2020, pp. 546--562.

\bibitem{liu2024infmae}
F.~Liu, C.~Gao, Y.~Zhang, J.~Guo, J.~Wang, and D.~Meng, ``Infmae: A foundation model in infrared modality,'' \emph{arXiv preprint arXiv:2402.00407}, 2024.

\bibitem{zhou2023position}
H.~Zhou, C.~Tian, Z.~Zhang, C.~Li, Y.~Ding, Y.~Xie, and Z.~Li, ``Position-aware relation learning for rgb-thermal salient object detection,'' \emph{IEEE Transactions on Image Processing}, vol.~32, pp. 2593--2607, 2023.

\bibitem{zhang2021weakly}
L.~Zhang, Z.~Liu, X.~Zhu, Z.~Song, X.~Yang, Z.~Lei, and H.~Qiao, ``Weakly aligned feature fusion for multimodal object detection,'' \emph{IEEE Transactions on Neural Networks and Learning Systems}, 2021.

\bibitem{tu2022weakly}
Z.~Tu, Z.~Li, C.~Li, and J.~Tang, ``Weakly alignment-free rgbt salient object detection with deep correlation network,'' \emph{IEEE Transactions on Image Processing}, vol.~31, pp. 3752--3764, 2022.

\bibitem{fu2023lraf}
H.~Fu, S.~Wang, P.~Duan, C.~Xiao, R.~Dian, S.~Li, and Z.~Li, ``Lraf-net: Long-range attention fusion network for visible--infrared object detection,'' \emph{IEEE Transactions on Neural Networks and Learning Systems}, 2023.

\bibitem{li2023stabilizing}
Q.~Li, C.~Zhang, Q.~Hu, P.~Zhu, H.~Fu, and L.~Chen, ``Stabilizing multispectral pedestrian detection with evidential hybrid fusion,'' \emph{IEEE Transactions on Circuits and Systems for Video Technology}, 2023.

\bibitem{shen2024icafusion}
J.~Shen, Y.~Chen, Y.~Liu, X.~Zuo, H.~Fan, and W.~Yang, ``Icafusion: Iterative cross-attention guided feature fusion for multispectral object detection,'' \emph{Pattern Recognition}, vol. 145, p. 109913, 2024.

\bibitem{yuan2024improving}
M.~Yuan, X.~Shi, N.~Wang, Y.~Wang, and X.~Wei, ``Improving rgb-infrared object detection with cascade alignment-guided transformer,'' \emph{Information Fusion}, vol. 105, p. 102246, 2024.

\bibitem{yuan2024c}
M.~Yuan and X.~Wei, ``C 2 former: Calibrated and complementary transformer for rgb-infrared object detection,'' \emph{IEEE Transactions on Geoscience and Remote Sensing}, 2024.

\bibitem{zhu2023multi}
Y.~Zhu, X.~Sun, M.~Wang, and H.~Huang, ``Multi-modal feature pyramid transformer for rgb-infrared object detection,'' \emph{IEEE Transactions on Intelligent Transportation Systems}, vol.~24, no.~9, pp. 9984--9995, 2023.

\bibitem{zhang2019weakly}
L.~Zhang, X.~Zhu, X.~Chen, X.~Yang, Z.~Lei, and Z.~Liu, ``Weakly aligned cross-modal learning for multispectral pedestrian detection,'' in \emph{Proceedings of the IEEE/CVF international conference on computer vision}, 2019, pp. 5127--5137.

\bibitem{zhou2020improving}
K.~Zhou, L.~Chen, and X.~Cao, ``Improving multispectral pedestrian detection by addressing modality imbalance problems,'' in \emph{Computer Vision--ECCV 2020: 16th European Conference, Glasgow, UK, August 23--28, 2020, Proceedings, Part XVIII 16}.\hskip 1em plus 0.5em minus 0.4em\relax Springer, 2020, pp. 787--803.

\bibitem{yuan2022translation}
M.~Yuan, Y.~Wang, and X.~Wei, ``Translation, scale and rotation: cross-modal alignment meets rgb-infrared vehicle detection,'' in \emph{European Conference on Computer Vision}.\hskip 1em plus 0.5em minus 0.4em\relax Springer, 2022, pp. 509--525.

\bibitem{redmon2016you}
J.~Redmon, S.~Divvala, R.~Girshick, and A.~Farhadi, ``You only look once: Unified, real-time object detection,'' in \emph{Proceedings of the IEEE conference on computer vision and pattern recognition}, 2016, pp. 779--788.

\bibitem{redmon2017yolo9000}
J.~Redmon and A.~Farhadi, ``Yolo9000: better, faster, stronger,'' in \emph{Proceedings of the IEEE conference on computer vision and pattern recognition}, 2017, pp. 7263--7271.

\bibitem{wagner2016multispectral}
J.~Wagner, V.~Fischer, M.~Herman, S.~Behnke \emph{et~al.}, ``Multispectral pedestrian detection using deep fusion convolutional neural networks.'' in \emph{ESANN}, vol. 587, 2016, pp. 509--514.

\bibitem{konig2017fully}
D.~Konig, M.~Adam, C.~Jarvers, G.~Layher, H.~Neumann, and M.~Teutsch, ``Fully convolutional region proposal networks for multispectral person detection,'' in \emph{Proceedings of the IEEE conference on computer vision and pattern recognition workshops}, 2017, pp. 49--56.

\bibitem{liu2016multispectral}
J.~Liu, S.~Zhang, S.~Wang, and D.~N. Metaxas, ``Multispectral deep neural networks for pedestrian detection,'' in \emph{27th British Machine Vision Conference, BMVC 2016}, 2016.

\bibitem{cao2023multimodal}
Y.~Cao, J.~Bin, J.~Hamari, E.~Blasch, and Z.~Liu, ``Multimodal object detection by channel switching and spatial attention,'' in \emph{Proceedings of the IEEE/CVF Conference on Computer Vision and Pattern Recognition}, 2023, pp. 403--411.

\bibitem{qingyun2022cross}
F.~Qingyun and W.~Zhaokui, ``Cross-modality attentive feature fusion for object detection in multispectral remote sensing imagery,'' \emph{Pattern Recognition}, vol. 130, p. 108786, 2022.

\bibitem{roszyk2022adopting}
K.~Roszyk, M.~R. Nowicki, and P.~Skrzypczy{\'n}ski, ``Adopting the yolov4 architecture for low-latency multispectral pedestrian detection in autonomous driving,'' \emph{Sensors}, vol.~22, no.~3, p. 1082, 2022.

\bibitem{qingyun2021cross}
F.~Qingyun, H.~Dapeng, and W.~Zhaokui, ``Cross-modality fusion transformer for multispectral object detection,'' \emph{arXiv preprint arXiv:2111.00273}, 2021.

\bibitem{li2019illumination}
C.~Li, D.~Song, R.~Tong, and M.~Tang, ``Illumination-aware faster r-cnn for robust multispectral pedestrian detection,'' \emph{Pattern Recognition}, vol.~85, pp. 161--171, 2019.

\bibitem{yang2022baanet}
X.~Yang, Y.~Qian, H.~Zhu, C.~Wang, and M.~Yang, ``Baanet: Learning bi-directional adaptive attention gates for multispectral pedestrian detection,'' in \emph{2022 international conference on robotics and automation (ICRA)}.\hskip 1em plus 0.5em minus 0.4em\relax IEEE, 2022, pp. 2920--2926.

\bibitem{li2018multispectral}
C.~Li, D.~Song, R.~Tong, and M.~Tang, ``Multispectral pedestrian detection via simultaneous detection and segmentation,'' in \emph{29th British Machine Vision Conference, BMVC 2018}.

\bibitem{cao2019box}
Y.~Cao, D.~Guan, Y.~Wu, J.~Yang, Y.~Cao, and M.~Y. Yang, ``Box-level segmentation supervised deep neural networks for accurate and real-time multispectral pedestrian detection,'' \emph{ISPRS journal of photogrammetry and remote sensing}, vol. 150, pp. 70--79, 2019.

\bibitem{zhang2020multispectral}
H.~Zhang, E.~Fromont, S.~Lefevre, and B.~Avignon, ``Multispectral fusion for object detection with cyclic fuse-and-refine blocks,'' in \emph{2020 IEEE International conference on image processing (ICIP)}.\hskip 1em plus 0.5em minus 0.4em\relax IEEE, 2020, pp. 276--280.

\bibitem{zhang2021guided}
H.~Zhang, E.~Fromont, S.~Lef{\`e}vre, and B.~Avignon, ``Guided attentive feature fusion for multispectral pedestrian detection,'' in \emph{Proceedings of the IEEE/CVF winter conference on applications of computer vision}, 2021, pp. 72--80.

\bibitem{kim2021uncertainty}
J.~U. Kim, S.~Park, and Y.~M. Ro, ``Uncertainty-guided cross-modal learning for robust multispectral pedestrian detection,'' \emph{IEEE Transactions on Circuits and Systems for Video Technology}, vol.~32, no.~3, pp. 1510--1523, 2021.

\bibitem{li2022confidence}
Q.~Li, C.~Zhang, Q.~Hu, H.~Fu, and P.~Zhu, ``Confidence-aware fusion using dempster-shafer theory for multispectral pedestrian detection,'' \emph{IEEE Transactions on Multimedia}, vol.~25, pp. 3420--3431, 2022.

\bibitem{carion2020end}
N.~Carion, F.~Massa, G.~Synnaeve, N.~Usunier, A.~Kirillov, and S.~Zagoruyko, ``End-to-end object detection with transformers,'' in \emph{European conference on computer vision}.\hskip 1em plus 0.5em minus 0.4em\relax Springer, 2020, pp. 213--229.

\bibitem{zhu2020deformable}
X.~Zhu, W.~Su, L.~Lu, B.~Li, X.~Wang, and J.~Dai, ``Deformable detr: Deformable transformers for end-to-end object detection,'' \emph{arXiv preprint arXiv:2010.04159}, 2020.

\bibitem{meng2021conditional}
D.~Meng, X.~Chen, Z.~Fan, G.~Zeng, H.~Li, Y.~Yuan, L.~Sun, and J.~Wang, ``Conditional detr for fast training convergence,'' in \emph{Proceedings of the IEEE/CVF international conference on computer vision}, 2021, pp. 3651--3660.

\bibitem{yao2021efficient}
Z.~Yao, J.~Ai, B.~Li, and C.~Zhang, ``Efficient detr: improving end-to-end object detector with dense prior,'' \emph{arXiv preprint arXiv:2104.01318}, 2021.

\bibitem{liu2022dab}
S.~Liu, F.~Li, H.~Zhang, X.~Yang, X.~Qi, H.~Su, J.~Zhu, and L.~Zhang, ``Dab-detr: Dynamic anchor boxes are better queries for detr,'' \emph{arXiv preprint arXiv:2201.12329}, 2022.

\bibitem{li2022dn}
F.~Li, H.~Zhang, S.~Liu, J.~Guo, L.~M. Ni, and L.~Zhang, ``Dn-detr: Accelerate detr training by introducing query denoising,'' in \emph{Proceedings of the IEEE/CVF conference on computer vision and pattern recognition}, 2022, pp. 13\,619--13\,627.

\bibitem{zhangdino}
H.~Zhang, F.~Li, S.~Liu, L.~Zhang, H.~Su, J.~Zhu, L.~Ni, and H.-Y. Shum, ``Dino: Detr with improved denoising anchor boxes for end-to-end object detection,'' in \emph{The Eleventh International Conference on Learning Representations}.

\bibitem{zhao2024detrs}
Y.~Zhao, W.~Lv, S.~Xu, J.~Wei, G.~Wang, Q.~Dang, Y.~Liu, and J.~Chen, ``Detrs beat yolos on real-time object detection,'' in \emph{Proceedings of the IEEE/CVF Conference on Computer Vision and Pattern Recognition}, 2024, pp. 16\,965--16\,974.

\bibitem{ma2021oriented}
T.~Ma, M.~Mao, H.~Zheng, P.~Gao, X.~Wang, S.~Han, E.~Ding, B.~Zhang, and D.~Doermann, ``Oriented object detection with transformer,'' \emph{arXiv preprint arXiv:2106.03146}, 2021.

\bibitem{dai2022ao2}
L.~Dai, H.~Liu, H.~Tang, Z.~Wu, and P.~Song, ``Ao2-detr: Arbitrary-oriented object detection transformer,'' \emph{IEEE Transactions on Circuits and Systems for Video Technology}, vol.~33, no.~5, pp. 2342--2356, 2022.

\bibitem{zeng2024ars}
Y.~Zeng, Y.~Chen, X.~Yang, Q.~Li, and J.~Yan, ``Ars-detr: Aspect ratio-sensitive detection transformer for aerial oriented object detection,'' \emph{IEEE Transactions on Geoscience and Remote Sensing}, vol.~62, pp. 1--15, 2024.

\bibitem{guo2024damsdet}
J.~Guo, C.~Gao, F.~Liu, D.~Meng, and X.~Gao, ``Damsdet: Dynamic adaptive multispectral detection transformer with competitive query selection and adaptive feature fusion,'' \emph{arXiv e-prints}, pp. arXiv--2403, 2024.

\bibitem{he2016deep}
K.~He, X.~Zhang, S.~Ren, and J.~Sun, ``Deep residual learning for image recognition,'' in \emph{Proceedings of the IEEE conference on computer vision and pattern recognition}, 2016, pp. 770--778.

\bibitem{zhang2023decoupled}
M.~Zhang, G.~Song, Y.~Liu, and H.~Li, ``Decoupled detr: Spatially disentangling localization and classification for improved end-to-end object detection,'' in \emph{Proceedings of the IEEE/CVF International Conference on Computer Vision}, 2023, pp. 6601--6610.

\bibitem{lin2017focal}
T.-Y. Lin, P.~Goyal, R.~Girshick, K.~He, and P.~Doll{\'a}r, ``Focal loss for dense object detection,'' in \emph{Proceedings of the IEEE international conference on computer vision}, 2017, pp. 2980--2988.

\bibitem{chen2020piou}
Z.~Chen, K.~Chen, W.~Lin, J.~See, H.~Yu, Y.~Ke, and C.~Yang, ``Piou loss: Towards accurate oriented object detection in complex environments,'' in \emph{Computer Vision--ECCV 2020: 16th European Conference, Glasgow, UK, August 23--28, 2020, Proceedings, Part V 16}.\hskip 1em plus 0.5em minus 0.4em\relax Springer, 2020, pp. 195--211.

\bibitem{han2021align}
J.~Han, J.~Ding, J.~Li, and G.-S. Xia, ``Align deep features for oriented object detection,'' \emph{IEEE transactions on geoscience and remote sensing}, vol.~60, pp. 1--11, 2021.

\bibitem{xie2021oriented}
X.~Xie, G.~Cheng, J.~Wang, X.~Yao, and J.~Han, ``Oriented r-cnn for object detection,'' in \emph{Proceedings of the IEEE/CVF international conference on computer vision}, 2021, pp. 3520--3529.

\bibitem{ding2019learning}
J.~Ding, N.~Xue, Y.~Long, G.-S. Xia, and Q.~Lu, ``Learning roi transformer for oriented object detection in aerial images,'' in \emph{Proceedings of the IEEE/CVF conference on computer vision and pattern recognition}, 2019, pp. 2849--2858.

\bibitem{li2022oriented}
W.~Li, Y.~Chen, K.~Hu, and J.~Zhu, ``Oriented reppoints for aerial object detection,'' in \emph{Proceedings of the IEEE/CVF conference on computer vision and pattern recognition}, 2022, pp. 1829--1838.

\bibitem{zhang2019cross}
L.~Zhang, Z.~Liu, S.~Zhang, X.~Yang, H.~Qiao, K.~Huang, and A.~Hussain, ``Cross-modality interactive attention network for multispectral pedestrian detection,'' \emph{Information Fusion}, vol.~50, pp. 20--29, 2019.

\bibitem{sun2022drone}
Y.~Sun, B.~Cao, P.~Zhu, and Q.~Hu, ``Drone-based rgb-infrared cross-modality vehicle detection via uncertainty-aware learning,'' \emph{IEEE Transactions on Circuits and Systems for Video Technology}, vol.~32, no.~10, pp. 6700--6713, 2022.

\bibitem{hwang2015multispectral}
S.~Hwang, J.~Park, N.~Kim, Y.~Choi, and I.~So~Kweon, ``Multispectral pedestrian detection: Benchmark dataset and baseline,'' in \emph{Proceedings of the IEEE conference on computer vision and pattern recognition}, 2015, pp. 1037--1045.

\bibitem{guan2019fusion}
D.~Guan, Y.~Cao, J.~Yang, Y.~Cao, and M.~Y. Yang, ``Fusion of multispectral data through illumination-aware deep neural networks for pedestrian detection,'' \emph{Information Fusion}, vol.~50, pp. 148--157, 2019.

\end{thebibliography}


\end{document}